\documentclass[sigconf]{acmart}

\usepackage{booktabs} 
\usepackage{braket}
\usepackage{amsmath}
\usepackage[toc,page]{appendix}
\usepackage{url}
\usepackage[utf8]{inputenc}

\usepackage{color,xcolor}
\usepackage{array}

\newcolumntype{N}{@{}m{0pt}@{}}

\hypersetup{draft}

\makeatletter
\newif\if@restonecol
\makeatother

\usepackage[ruled,vlined]{algorithm2e}

\usepackage{amssymb,amsmath}
\usepackage{booktabs}
\usepackage{mathrsfs}
\usepackage{makecell}

\copyrightyear{2019}
\acmYear{2019} 
\setcopyright{iw3c2w3}
\acmConference[WWW '19]{Proceedings of the 2019 World Wide Web Conference}{May 13--17, 2019}{San Francisco, CA, USA}
\acmBooktitle{Proceedings of the 2019 World Wide Web Conference (WWW '19), May 13--17, 2019, San Francisco, CA, USA}
\acmPrice{}
\acmDOI{10.1145/3308558.3313516}
\acmISBN{978-1-4503-6674-8/19/05}

\begin{document}
\large

\title{Semantic Hilbert Space for Text Representation Learning}

\author{Benyou Wang,~Qiuchi Li,~Massimo Melucci}
\affiliation{%
  \institution{University of Padua}
  \streetaddress{Via Giovanni Gradenigo, 6/B}
  \city{Padua}
  \state{Italy}
  \postcode{PD 35121}
}

\email{wang,qiuchili,melo@dei.unipd.it}

\author{Dawei Song}
\affiliation{%
  \institution{Beijing Institute of Technology}
  \streetaddress{Haidian}
  \city{Beijing}
  \state{China}
  \postcode{100081}
}
\email{dawei.song2010@gmail.com}
\thanks{$^\dag$ Benyou Wang and Qiuchi Li contribute equally and share the co-first authorship. \\
*Massimo Melucci (melo@dei.unipd.it) is the corresponding author.}


%
%
%
%
%
%
\renewcommand{\shortauthors}{B. Wang et al.}

\begin{abstract}
Capturing the meaning of sentences has long been a challenging task. Current models tend to apply linear combinations of word features to conduct semantic composition for bigger-granularity units e.g. phrases, sentences, and documents. However, the semantic linearity does not always hold in human language. For instance, the meaning of the phrase ``ivory tower'' cannot be deduced by linearly combining the meanings of ``ivory'' and ``tower''. 
To address this issue,  we propose a new framework that models different levels of semantic units (e.g. sememe, word, sentence, and semantic abstraction) on a single \textit{Semantic Hilbert Space}, which naturally admits a non-linear semantic composition by means of a complex-valued vector word representation. 
An end-to-end neural network~\footnote{https://github.com/wabyking/qnn} is proposed to implement the framework in the text classification task, and evaluation results on six benchmarking text classification datasets demonstrate the effectiveness, robustness and self-explanation power of the proposed model. Furthermore, intuitive case studies are conducted to help end users to understand how the framework works.


\end{abstract}



\maketitle

\section{Introduction}
\sloppypar

In natural language understanding, it is crucial, yet challenging, to model sentences and capture their meanings. Essentially, most statistical machine learning models~\cite{hill_learning_2016_a,Kiros:2015:SV:2969442.2969607,arora_simple_2016,D17-1070,pagliardini_unsupervised_2017} are built within a linear bottom-up framework, where words are the basic features adopting a low-dimensional vector representation, and a sentence is modeled as a linear combination of individual word vectors. Such linear semantic composition is efficient, but does not always hold in human language. For example, the phrase ``ivory tower'', which means ``a state of privileged seclusion or separation from the facts and practicalities of the real world'', is not a linear combination of the individual meanings of ``ivory'' and ``tower''. Instead, it carries a new meaning. We are therefore motivated to investigate a new language modeling paradigm to account for such intricate non-linear combination of word meanings. 

Drawing inspiration from the recent findings in the emerging research area of quantum cognition, which suggest that human cognition~\cite{busemeyer2012quantum,aerts_concepts_2013,aerts_quantum_2014} especially language understanding~\cite{bruza2008entangling,bruza2009there,wang2016exploration} exhibit certain non-classical phenomena (i.e. quantum-like phenomena), we propose a theoretical framework, named~\textit{Semantic Hilbert Space}, to formulate quantum-like phenomena in language understanding and to model different levels of semantic units in a unified space.

In Semantic Hilbert Space, we assume that words can be modeled as microscopic particles in superposition states, over the basic sememes (i.e. minimum semantic units in linguistics), while a combination of word meanings can be viewed as a mixed system of particles. 
The Semantic Hilbert Space represents different levels of semantic units, ranging from basic sememes, words and sentences, on a unified complex-valued vector space. This is fundamentally different from existing quantum-inspired neural networks for question answering~\cite{zhang_end_to_end_2018,zhang2018quantum2} which are based on a real vector space. In addition, we introduce a new semantic abstraction, named as \textit{Semantic Measurements}, which are also embedded in the same vector space and trainable to extract high-level features from the mixed system. 

As shown in Fig.~\ref{fig:framework}, the Semantic Hilbert Space is built on the basis of quantum probability (QP), which is the probability theory for explaining the uncertainty of quantum superposition. As quantum superposition requires the use of the complex field, Semantic Hilbert Space has complex values and operators. In particular, the probability function is implemented by a unique (complex) density operator. 

Semantic Hilbert Space adopts a complex-valued vector representation of unit length, where each component adopts an amplitude-phase form $z = re^{i\phi}$. We hereby hypothesize that the amplitude $r$ and complex phase $\phi$ can be used to encode different levels of semantics such as lexical-level co-occurrence, hidden sentiment polarity or topic-level semantics. When word vectors are combined, even in a simple complex-valued addition form, the resulting expression will entail a non-linear composition of amplitudes and phases, thus indicating a complicated fusion of different levels of semantics. A more detailed explanation is given in Sec. 3. In this way, the complex-valued word embedding is fundamentally different from existing real-valued word embedding. A series of ablation tests indicate that the complex-valued word embedding can increase performance.

The Semantic Hilbert Space is an abstract representation of our approach to modeling language through QP. At the level of implementation, an efficient and effective computational framework is needed to cope with large text collections. To do so, we propose an end-to-end neural network architecture, which provides means for training of the network components. Each component corresponds to a physical meaning of quantum probability with well-defined mathematical constraints. Moreover, each component is easier to understand than the kernels in convolutional neural network and cells in recurrent neural networks.

The network proposed in this paper is evaluated on six benchmarking datasets for text classification and achieves a steady increase over existing models. Moreover, it is shown that the proposed network is advantageous due to its high robustness and self-explanation capability.

\section{Semantic Hilbert Space}
\label{semantic_hilbert_space}

\begin{figure*}[ht]\label{quantum}
  \includegraphics[width=\textwidth]{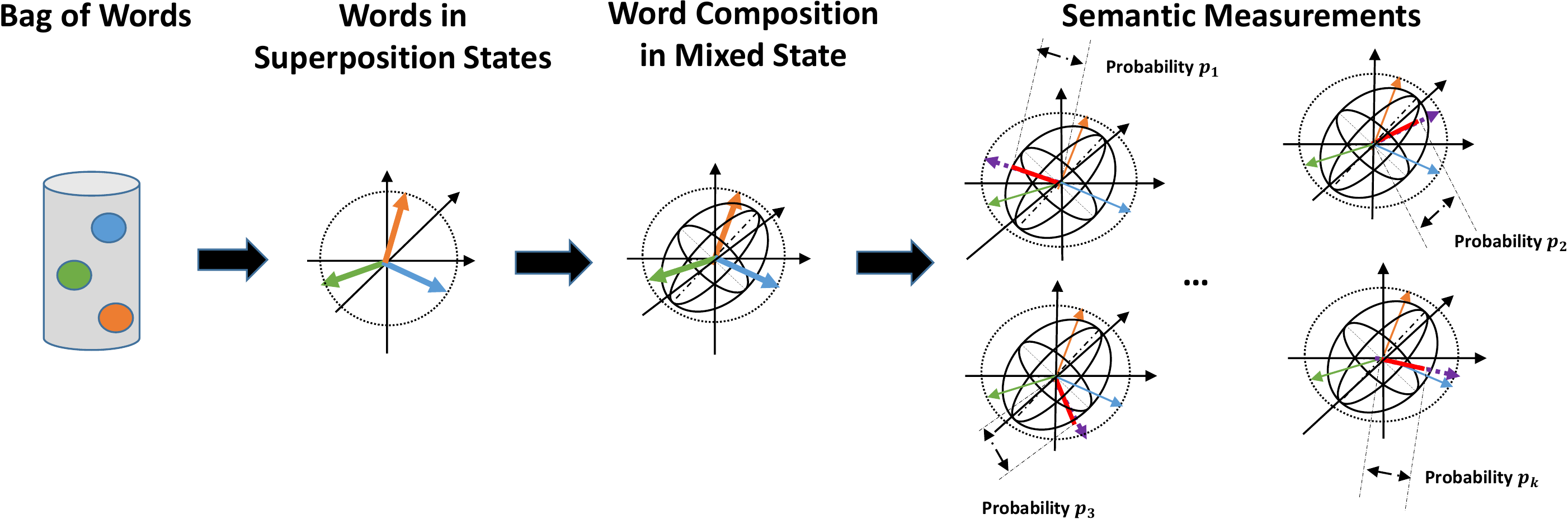}
  \caption{Illustration of Semantic Hilbert Space. The green, blue and orange
    colors correspond to three different words modeled as quantum particles. The
    black dotted circle represents the unit ball in the Semantic Hilbert Space.
    The ellipsoid in solid line refers to the quantum probability distribution
    defined by the density matrix of the word composition. The purple lines are
    semantic measurements. The intersections of the ellipsoids and semantic
    measurements are in thick red lines, the lengths of which correspond to
    measurement probabilities.} 
\label{fig:framework} 
\end{figure*}

The mathematical foundation of Quantum Theory is established on a Hilbert Space
over the complex field. In order to borrow the underlying mathematical formalism
of quantum theory for language understanding, it is necessary to build such a
Hilbert Space for language representation. In this study, we build a
\textit{Semantic Hilbert Space} $\mathcal{H}$ over the complex field. As is
illustrated in Fig.~\ref{fig:framework}, multiple levels of semantic units are
modeled on this common Semantic Hilbert Space. In the rest of this section, the
semantic units under modeling are introduced separately. 

We follow the standard \textit{Dirac Notation} for Quantum Theory. A unit vector
and its transpose are denoted as a ket $\ket{\mu}$ and a bra $\bra{\mu}$,
respectively. The inner product and outer product of two unit vectors $\vec{u}$
and $\vec{v}$ are denoted as $\braket{u \vert v }$ and $\ket{u}\bra{v}$
respectively.

\subsection{Sememes}

Sememes are the minimal non-separable semantic units of word meanings in
language universals~\cite{goddard1994semantic}. For example, the word
``blacksmith'' is composed of sememes ``human'', ``occupation'', ``metal'' and
``industrial''. We assume that the Semantic Hilbert Space $\mathcal{H}$ is
spanned by a set of orthogonal basis $\{\ket{e_j}\}_{j=1}^n$ corresponding to a
finite closed set of sememes $\{e_j\}_{j=1}^n$. In the quantum language, the set
of sememes are modeled as \textit{basis states}, which is the basis for
representing any quantum state. In Fig.~\ref{fig:framework}, the axes of the
Semantic Hilbert Space correspond to the set of sememe states, and semantic
units with larger granularity are represented based on quantum probability.

\subsection{Words}

The meaning of a word is a combination of sememes. We adopt the concept of
\textit{superposition} to formulate this combination. Essentially, a word $w$ is
modeled as a quantum particle in \textit{superposition state}, represented by a
unit-length vector in the Semantic Hilbert Space $\mathcal{H}$, as can be seen
in Fig.~\ref{fig:framework}. It can be written as a linear combination of the
basis states for sememes:
\begin{equation}
  \label{superposition}
  \ket{w}  =  \sum_{j=1}^n r_j e^{i\phi_j} \ket{ e_j }
\end{equation}
where the complex-valued weight $r_j e^{i\phi_j}$ denotes how much the meaning
of word $w$ is associated with the sememe $e_j$. Here $\{r_j\}_{j=1}^n$ are
non-negative real-valued amplitudes satisfying $\sum_{j=1}^n {r_j}^2$ =1 and
$\phi_j \in [-\pi,\pi]$ are the corresponding complex phases. We could also
transfer the complex number in a complex plane as $re^{i\phi} = r\cos\phi +
ir\sin\phi$.

It is worth noting that the complex phases $\{\phi_j\}$ are crucial as they
implicitly entail the \textit{quantum interference} between words. Suppose two
words $w_1$ and $w_2$ are associated to weights $r_j^{(1)} e^{i\phi_j^{(1)}}$ and
$r_j^{(2)} e^{i\phi_j^{(2)}}$ for the sememe $e_j$. The two words in combination
are therefore at the state $e_j$ with a probability of
\begin{equation}
  \left|r_j^{(1)} e^{i\phi_j^{(1)}}+r_j^{(2)} e^{i\phi_j^{(2)}}\right|^2 
  =
  \left|r_j^{(1)}\right|^2 + \left|r_j^{(2)}\right|^2 +
  2r_j^{(1)}r_j^{(2)}\cos\left(\phi_j^{(1)}-\phi_j^{(2)}\right) 
\end{equation}
where the term $2r_j^{(1)}r_j^{(2)}\cos(\phi_j^{(1)}-\phi_j^{(2)})$ reflects the
interference between the two words, where as the classical case corresponds to a
particular case $\phi_j^{(1)} = \phi_j^{(2)} = 0$.

\subsection{Semantic Compositions}
\label{mixed_system}

As is illustrated in Fig.~\ref{fig:framework}, we view a word composition (e.g.
a sentence) as a bag of words~\cite{harris1954distributional}, each of which is
modeled as a particle in superposition state on the Semantic Hilbert Space
$\mathcal{H}$. To obtain the semantic composition of words, we leverage the
concept of \textit{quantum mixture} and formulate the word composition as a mixed
system composed of the word superposition states. The system is in a
\textit{mixed state} represented by a $n$-by-$n$ density matrix $\rho$ on
$\mathcal{H}$, which is positive semi-definite with trace 1. It is computed as
follows:
\begin{equation}
  \label{mixture}
  \rho = \sum_i { p(i) \ket{w_i} \bra{w_i}},
\end{equation}
where $\ket{w_i}$ denotes the superposition state of the $i$-th word and $p(i)$
is the classical probability of the state $\ket{w_i}$ with $\sum_i p(i) = 1$. It
determines the contribution of the word $w_i$ to the overall semantics.

The complex-valued density matrix $\rho$ can be seen non-classical distribution
of sememes in $\mathcal{H}$. Its diagonal elements are real and form a classical
distribution of sememes, while its complex-valued off--diagonal entries encode
the interplay between sememes, which in turn gives rise to the interference
between words. A density matrix assigns a probability value for any state on
$\mathcal{H}$ such that the values for any set of orthogonal states sum up to
1~\cite{gleason1957measures}. Hence it is visualized as an ellipsoid in
Fig.~\ref{fig:framework}, assigning a quantum probability to a unit vector with the  intersection length.

\subsection{Semantic Measurements}

As a non-classical probability distribution, a sentence density matrix carries
rich information and in particular it contains all the information about a
quantum system. In order to extract the relevant information to a concrete task
from the semantic composition, we build a set of measurements and compute the
probability that the mixed system falls onto each of the measurements as a
high-level abstraction of the semantic composition.

Suppose our proposed \textit{semantic measurements} are associated with a set of measurement projectors $\{P_i\}_{i=1}^k$. According to the Born's rule~\cite{born_zur_1926}, applying the measurement projector $P_i$ onto the sentence density matrix $\rho$ yields the following result: 
\begin{equation} 
  \label{born}
 p_i = tr(P_i\rho)
\end{equation}
Here, we only consider pure states as measurement states, i.e. $P_i =
\ket{v_i}\bra{v_i}$. Moreover, we ignore the constraints of the measurements
states $\{\ket{v_i}\}_{i=1}^k$ (i.e. orthogonality or completeness), but keep
them trainable, so that the most suitable measurements can be determined
automatically by the data in a concrete task, such as classification or
regression. In this way, the trainable semantic measurements can be understood
as a similar approach to supervised dimensionality
reduction~\cite{fisher_use_1936}, but in a quantum probability framework with
complex values.

\section{Quantum Probability Driven Network}

\begin{figure*}[t]
  \includegraphics[width=\textwidth]{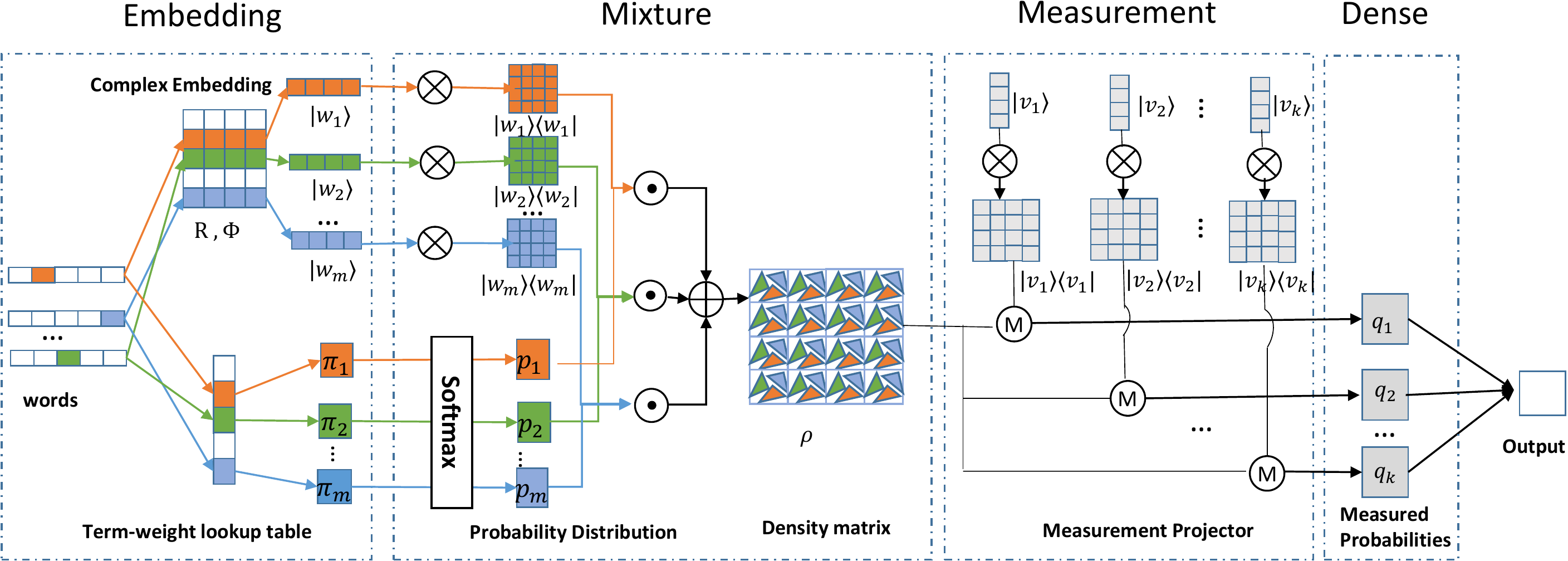}
  \caption{Architecture of Quantum probability-driven Neural Network. $\bigodot$ means that a matrix multiplies a number with each elements.  $\bigoplus$ refers to a element-wise addition.  $\bigotimes$ denotes a outer production to a vector, $\textcircled{m} $ means a measurement operation according to Eq.~\ref{born}.}
\label{fig:architecture1}
\end{figure*} 

In order to implement the proposed framework, we further propose an end-to-end
neural network  based on quantum probability.
Fig. \ref{fig:architecture1} shows the architecture of the proposed Quantum
Probability Driven Network (QPDN). 
The embedding layer, which is composed of a unit complex-valued embedding and a
term-weight lookup table, captures the basic lexical features. The
mixture layer is designed to combine the low-level bag-of-word features with
an additive complex-valued outer product operation. The measurement layer
adopts a set of trainable semantic measurements to extract the higher-level
features for the final linear classifier. In the following we will introduce the
architecture layer by layer.

\subsection{Embedding Layer}

The parameters of the embedding layer are $\{R, \Phi, \Pi\}$, respectively,
denoting the amplitude embedding, the phase embedding, and the term-weight lookup
table. Eq.~\ref{superposition} expresses a quantum representation as a
unit-length, complex-valued vector representation for a word $w$, i.e.
$\ket{w} = [r_1 e^{i\phi_1}, r_2 e^{i\phi_2} ... r_n e^{i\phi_n} ]^T$. The term-weight lookup table is used to weight words for semantic
combinations, which will be described in the next subsection.
During training, word embeddings need to be normalized to unit length after each
batch. 

This representation allows for a non-linear composition of amplitudes and phases in its mathematical form. Suppose two words $w_1$ and $w_2$ are of weights $r_j^{(1)} e^{i\phi_j^{(1)}}$ and $r_j^{(2)} e^{i\phi_j^{(2)}}$ for the $j^{th}$ dimension (corresponding to the $j^{th}$ sememe). The combination of $w_1$ and $w_2$ gives rise to a weight $r_je^{i\phi_j}$ for the $j^{th}$ dimension computed as
\begin{equation}
\begin{aligned}
    r_je^{i\phi_j} & = r_j^{(1)} e^{i\phi_j^{(1)}}+r_j^{(2)} e^{i\phi_j^{(2)}} \\
    & = \sqrt{|r_j^{(1)}|^2 + |r_j^{(2)}|^2 +
  2r_j^{(1)}r_j^{(2)}cos(\phi_j^{(1)}-\phi_j^{(2)})} \\ &\times  
e^{i\arctan \left( \frac{ r_j^{(1)}\sin(\phi_j^{(1)}) + r_j^{(2)}\sin(\phi_j^{(2)})}{ r_j^{(1)}\cos(\phi_j^{(1)})+r_j^{(2)}\cos(\phi_j^{(2)})} \right) }
\end{aligned}
\end{equation}
Where both $r_j$ and $\phi_j$ is a non-linear combination of $r_j^{(1)}$,$r_j^{(2)}$,$\phi_j^{(1)}$ and $\phi_j^{(2)}$. If the amplitudes and phases are associated to different levels of information, the amplitude-phase representation then naturally gives rise to a non-linear fusion of information.

\subsection{Mixture Layer}

A sentence is modeled as a density matrix, which is constructed in Sec.~\ref{mixed_system}. Instead of using uniform weights in
Eq.~\ref{mixture}, word-sensitive weights are used for each word, which is
commonly used in IR, e.g. inverse document frequency (IDF) as a word-dependent
weight in TF-IDF scheme~\cite{sparck1972statistical}. 


In order to guarantee the unit trace length for density matrix, the word weights
which are from the lookup table in a sentence are normalized to a probability
value through a softmax operation: $p(i) ={e^{\pi(w_i)}}\,/\,{\sum^m_j e^{\pi(w_j)}}.$ Compared to the IDF weight, the
normalized weight for a specific word in our approach is not static but updated
adaptively in the training phase. Even in the inference/test phase, the real term
weight i.e. $p(w_i)$ is also not static, but highly depends on the neighbor
context words through nonlinear softmax function.


\subsection{Measurement Layer}

The measurement layer adopts a set of one-rank measurement projectors $\{\ket{v_i}\bra{v_i} \}_{i=1}^{k}$ where $\ket{v_i}\bra{v_i}$ is the outer product of its corresponding state in Semantic Hilbert Space $\ket{v_i}$. After each measurement, we can obtain one probability for each measurement state like $q_j= tr(\rho \ket{v_j} \bra{v_j} )$. Finally, we can obtain a vector $\vec{q}=[q_1,q_2,...q_k]$. Similarly to the word vectors, the states $\ket{v_i}$ are represented as unit states and normalized after several batches.

\subsection{Dense Layer}

The vector $\vec{q}$ in the measurement layer consists of $k$ positive scalar
numbers and it is used to infer the label for a given sentence. A dense layer
with softmax activation is adopted after the measurement layer to get a
classification probability distribution, i.e. $\widehat{\vec{y}}=
\mbox{softmax}(\vec{q} \cdot W )$. The loss is designed as a cross-entropy loss
between $\widehat{\vec{y}}$ and the one-hot label $\vec{y}$.

\section{Experiments}

\begin{table}[t]
  \begin{center}
    \small
    \begin{tabular}{l|lllll}
      \hline
      Dataset & train & test &  vocab. & task & Classes\\ \hline
      CR & 4K& CV  & 6K & product reviews & 2 \\
      MPQA & 11k & CV  &6K & opinion polarity & 2\\ 
      SUBJ & 10k &CV  &21k& subjectivity & 2 \\
      MR & 11.9k &CV  &20k& movie reviews & 2  \\ 
      SST & 67k &2.2k  &18k& movie reviews & 2 \\ 
      TREC & 5.4k &0.5k  &10k& Question & 6\\ 
      \hline
    \end{tabular}
  \end{center}
  \caption{Dataset Statistics. (CV means 10-fold cross validation for testing performance.)}\label{table:1}
  \label{tab:1}
  \vspace{-10pt}
\end{table}
Our model was evaluated on six datasets for text classification: CR customer
review~\cite{hu_mining_2004}, MPQA opinion
polarity~\cite{wiebe_annotating_2005}, SUBJ sentence
subjectivity~\cite{pang_seeing_2005}, MR movie review~\cite{pang_seeing_2005},
SST binary sentiment classification~\cite{socher_recursive_2013}, and TREC
question classification~\cite{li_learning_2002}. The statistics of them are
shown in Tab. \ref{tab:1}.

We compared the proposed QPDN with various models, including Uni-TFIDF,
Word2vec, FastText~\cite{joulin2016bag} and
Sent2Vec~\cite{pagliardini_unsupervised_2018} as unsupervised representation
learning baselines, CaptionRep~\cite{hill_learning_2016_a} and
DictRep~\cite{hill_learning_2016_b} as supervised representation learning
baselines, as well as CNN~\cite{kim_convolutional_2014} and
BiLSTM~\cite{conneau_supervised_2017} for advanced deep neural networks. In Tab.~\ref{table:2}, we
reported the classification accuracy values of these models from the original
papers.

We used Glove word vectors~\cite{pennington2014glove} with 50,100,200 and 300
dimensions respectively. The amplitude embedding values are initialized by
L2-norm, while the phases in complex-valued embedding are randomly initialized
in $-\pi$ to $\pi$.  We searched for the best performance in a parameter pool,
which contains a learning rate in $\{1\text{E-}3,1\text{E-}4,1\text{E-}5,1\text{E-}6\}$, an
L2-regularization ratio in $\{1\text{E-}5,1\text{E-}6,1\text{E-}7,1\text{E-}8\}$, a batch size in
$\{8,16,32,64,128\}$, and the number of measurements in
$\{5,10,20,50,100,200\}$.

The main parameters in our model are $R$ and $\Phi$. Since both of them are $n
\times |V|$ in shape, the number of parameters is roughly two times that of
FastText~\cite{mikolov2013efficient}. For the other parameters, $\Pi$ is $|V|
\times 1$, $\{\ket{v_i}\}_{i=1}^k$ is $k\times 2n$, while $W$ is $k \times |L|$
with $L$ being the label set. Apart from word embeddings, the model is robust
with limited scale at $k \times 2n + n \times |V| + k \times |L|$ for the number
of parameters.

\begin{table}[t]
\begin{center}
\small
\caption{Experimental Results in percentage (\%). The best performed value (except for CNN/LSTM) for each dataset is in bold. where $^\dag$ means a significant improvement over FasText.   }

\label{table:2}
\begin{tabular}{lcccccc}
 \hline
  \textbf{Model}& \textbf{CR} & \textbf{MPQA }& \textbf{MR} & \textbf{SST} & \textbf{SUBJ} & \textbf{TREC}\\ \hline
  Uni-TFIDF & 79.2& 82.4 & 73.7 & -  &90.3 &85.0\\ 
  Word2vec  & 79.8 & \textbf{88.3} & 77.7 & 79.7&90.9 &83.6\\
  FastText~\cite{joulin2016bag}  & 78.9& 87.4 & 76.5 & 78.8 &91.6 & 81.8\\ 
  Sent2Vec~\cite{pagliardini_unsupervised_2018} & 79.1 & 87.2 & 76.3 & 80.2& 91.2& 85.8 \\
  CaptionRep~\cite{hill_learning_2016_a}  & 69.3& 70.8 & 61.9 & - &77.4 & 72.2\\ 
  DictRep~\cite{hill_learning_2016_b}  & 78.7& 87.2 & 76.7 & - &90.7 & 81.0\\ \hline 

 Ours: QPDN & \textbf{81.0$^\dag$ }& 87.0 & \textbf{80.1$^\dag$} & \textbf{83.9$^\dag$}  &\textbf{92.7$^\dag$} & \textbf{88.2$^\dag$}\\ \hline
   CNN~\cite{kim_convolutional_2014}& 81.5& 89.4 & 81.1& 88.1&93.6 &92.4\\ 
  BiLSTM~\cite{conneau_supervised_2017} & 81.3& 88.7 & 77.5 & 80.7 & 89.6 &85.2\\ \hline
\end{tabular}
\end{center}
\vspace{-10pt}
\end{table}

The results in Tab.~\ref{table:2} demonstrate the effectiveness of our model and an improvement of  classification accuracies over some strong baseline supervised and
unsupervised representation models on most of the datasets except MPQA. In
comparison with more advanced models including BiLSTM and CNN, our model
generally performed better than BiLSTM with increased accuracy values on the
multi-class classification dataset (TREC) and three binary text classification
datasets (MR, SST \& SUBJ). However, it under-performed CNN on all 6 datasets
with a difference of over 2\% on 3 of them (MPQA, SST \& TREC), probably because
that it uses fewer parameters and simpler structures.

We argue that QPDN achieved a good balance between effectiveness and efficiency,
due to the fact that it outperforms BiLSTM.

\section{Discussions}
\label{discussion}

This section discusses the power of self-explanation and conducts an ablation test to examine the usefulness of important components of the network, especially the complex-valued word embedding.

\paragraph{Self-explanation Components}
\label{sec:Interpretability}

As is shown in Tab.~\ref{table:Interpretability}, all components in our model have a clear physical meaning corresponding to quantum probability, where classical Deep Neural Network (DNN) can not well explain the role each component plays in the network. Essentially, we constructed a bottom-up framework to represent each level of semantic units on a uniform Semantic Hilbert Space, from the minimum semantic unit, i.e. sememe, to the sentence representation. The framework was operationalized through superposition, mixture and semantic measurements. 
\begin{table}[t]
\begin{center}
\small
\caption{Physical meanings and constraints}\label{table:corresponding}
\begin{tabular}{lll}
 \hline
  Components &  DNN &  QPDN \\ \hline

Sememe &\makecell [l] { - }  &\makecell [l] {  basis vector / \textbf{basis state} \\ $ \{ w| w
 \in \mathcal{C}^n, ||w||_2 = 1,\} $ \\ complete~\&orthogonal  } \\
Word &\makecell [l] { real vector \\ $(-\infty, \infty)$}  &\makecell [l] { unit complex vector / \textbf{superposition state} \\ $ \{ w| w
 \in \mathcal{C}^n, ||w||_2 = 1 \}$  } \\ 
\makecell [l] {Low-level \\ representation} & \makecell [l] { real vector \\ $(-\infty, \infty)$}  &\makecell [l] { density matrix / \textbf{mixed system} \\ $ \{{\rho}| \rho =\rho ^*, tr(\rho) = 1 \} $  }\\ 
Abstraction &  \makecell [l] { CNN/RNN \\ $(-\infty, \infty)$}&  \makecell [l] { unit complex vector / \textbf{measurement} \\ $\{ w|w \in \mathcal{C}^n, ||w||_2 = 1 \}$  } \\
\makecell [l] {High-level \\representation} & \makecell [l] { real vector \\ $(-\infty, \infty)$}  &  \makecell [l] { probabilities/ \textbf{measured probability} \\ $(0,1)$} \\ 
 \hline
 \label{table:Interpretability}
\end{tabular}
\end{center}
\vspace{-10pt}
\end{table}

\paragraph{Ablation Test}

An ablation test was conducted to examine how  each component influences the final performance of QPDN. In particular, a double-length real word embedding network was implemented to examine the use of complex-valued word embedding, while mean weights and IDF weights were compared with our proposed trainable weights. A set of non-trainable orthogonal projectors and a dense layer on top of the sentence density matrix were implemented to analyze the effect of trainable semantic measurements.

Due to limited space, we only reported the ablation test result for SST, which is the largest and hence the most representative dataset.
We used 100-dimensional real-valued word vectors and 50-dimensional complex-valued vectors for the models in the ablation test. 
All models under ablation were comparable in terms of time cost.
Tab. ~\ref{table:Ablation} showed that each component plays an important role in the QPDN model. In particular, replacing complex embedding with double-dimension real word embedding led to a 5\% drop in performance, which indicates that the complex-valued word embedding was not merely doubling the number of parameters.

The comparison with IDF and mean weights showed that the data-driven scheme gave rise to high-quality word weights. The comparison with non-trainable projectors and directly applying a dense layer on the density matrix showed that trainable measurements bring benefits to the network.

\begin{table}[t]
\begin{center}
\small
\caption{Ablation Test}\label{table:Ablation}
\begin{tabular}{lll}
 \hline
Setting &SST  & $\Delta$\\ \hline
FastText~\cite{joulin2016bag} & 0.7880 & -0.0511\\  
FastText~\cite{joulin2016bag} with double-dimension real word vectors  &  0.7883&-0.0508\\ 
fixed amplitude part but trainable phase part & 0.8199 &-0.0192  \\  \hline
replace trainable weights with fixed mean weights & 0.8303 &-0.0088\\ 
replace trainable weights with  fixed IDF weights & 0.8259 &-0.0132\\  \hline
non-trainable projectors with fixed orthogonal ones & 0.8171 &-0.0220  \\   
replace projectors with dense layer & 0.8221 &-0.0170  \\  \hline
QPDN & \textbf{0.8391} & -  \\
 \hline
\end{tabular}
\end{center}
\vspace{-10pt}
\end{table}

\paragraph{Discriminative Semantic Directions}
\label{sec:cases}

In order to better understand the well-trained measurement projectors, we obtained the top 10 nearest words in the complex-valued vector space for each trained measurement state (like $\ket{v_i}$), using KD tree~\cite{bentley1975multidimensional}. 
Due to limited space, we selected five measurements from the trained model for the MR dataset, and selected words from the top 10 nearest words to each measurement. As can be seen in Tab.~\ref{tab:measurement_case_study}, the first measurement was roughly about time, the second one was related to verb words which mainly mean `motivating others'. The third measurement grouped uncommon non-English words together. The last two measurements also grouped words sharing similar meanings. It is therefore interesting to see that relevant words can somehow be grouped together into certain topics during the training process, which may be discriminative for the given task.

\begin{table}[t]

\begin{center}
\small
\caption{The learned measurement for dataset MR. They are selected according to nearest words for a measurement vector in Semantic Hibert Space}\label{tab:measurement_case_study}
\begin{tabular}{cl}
 \hline
 Measurement & Selected neighborhood words \\ \hline
1 & change, months, upscale, recently, aftermath  \\  \hline
2 &  compelled, promised, conspire, convince, trusting\\  \hline
3 &  goo, vez, errol, esperanza, ana  \\  \hline 
4 &  ice, heal, blessedly, sustains, make  \\   \hline
5 &  continue, warned, preposterousness, adding, falseness  \\  \hline
\end{tabular}
\end{center}
\vspace{-10pt}
\end{table}
\section{Conclusions}

In order to better model the non-linearity of word semantic composition, we have developed a quantum-inspired framework that models different granularities of semantic units on the same Semantic Hilbert Space, and implement this framework into an end-to-end text classification network. The network showed a promising performance on six benchmarking text datasets, in terms of effectiveness, robustness and self-explanation ability. Moreover, the complex-valued word embedding, which inherently achieved non-linear combination of word meanings, brought benefits to the classification accuracy in a comprehensive ablation study.

This work is among the first step to apply the quantum probabilistic framework to text modeling. We believe it is a promising direction. In the future, we would like to further extend this work by considering deeper and more complicated structures such as attention or memory mechanism in language, in order to investigate related quantum-like phenomena on textual data to provide more intuitive insights. Additionally, Semantic Hilbert Space in a tensor space is also worthy to be explored like \cite{zhang2018quantum2}, which may provide more interesting insights for current communities. 
\section*{ACKNOWLEDGEMENT}
This work is supported by the Quantum Access and Retrieval Theory (QUARTZ) project, which has received funding from the European Union's Horizon 2020 research and innovation programme under the Marie Skłodowska-Curie grant
agreement No. 721321, and by the National Key Research and Development Program of China (grant No. 2018YFC0831700), Natural Science Foundation of China (grant No. U1636203), and Major Project of Zhejiang Lab (grant No. 2019DH0ZX01).

\clearpage

\bibliographystyle{ACM-Reference-Format}
\bibliography{sample-bibliography}

\end{document}